\documentclass[letterpaper, 10 pt, conference]{ieeeconf}  

\IEEEoverridecommandlockouts                              

\usepackage{url}
\usepackage[ruled,vlined,linesnumbered,boxed,noend]{algorithm2e}
\usepackage[hidelinks]{hyperref}
\usepackage[utf8]{inputenc}
\usepackage[small]{caption}
\usepackage{graphicx}
\usepackage{wrapfig}
\usepackage{subcaption}
\usepackage{bm}
\usepackage{amsmath}
\usepackage{cleveref}
\usepackage{siunitx}
\usepackage{amsfonts,amssymb}
\usepackage{graphicx}
\usepackage{booktabs}
\usepackage{soul}
\usepackage{color}

\usepackage{balance}

\DeclareRobustCommand{\SURQ}{\mbox{Surrogate-Q} } 

\DeclareMathOperator*{\argmax}{arg\,max}

\title{\LARGE \bf
Deep Surrogate Q-Learning for Autonomous Driving
}

\author{Maria Kalweit$^{1}$, Gabriel Kalweit$^{1}$, Moritz Werling$^{2}$ and Joschka Boedecker$^{1, 3}$
\thanks{$^{1}$Neurorobotics Lab, University of Freiburg, Germany.
        {\tt\small {kalweitm,kalweitg,jboedeck}@cs.uni-freiburg.de}}%
\thanks{$^{2}$BMWGroup, Unterschleissheim, Germany.
        {\tt\small Moritz.Werling@bmw.de}}%
\thanks{$^{3}$BrainLinks-BrainTools, University of Freiburg, Germany.}%
}

\usepackage{tikz} 

\newcommand\copyrighttext{%
  \footnotesize \textcopyright 2022 IEEE. Personal use of this material is permitted. Permission from IEEE must be obtained for all other uses, in any current or future media, including reprinting/republishing this material for advertising or promotional purposes, creating new collective works, for resale or redistribution to servers or lists, or reuse of any copyrighted component of this work in other works.}

\newcommand\copyrightnotice{%
\begin{tikzpicture}[remember picture,overlay]
\node[anchor=south,yshift=15pt] at (current page.south) {\fbox{\parbox{\dimexpr\textwidth-\fboxsep-\fboxrule\relax}{\copyrighttext}}};
\end{tikzpicture}%
}

\begin{document}

\maketitle
\copyrightnotice
\thispagestyle{empty}
\pagestyle{empty}

\begin{abstract}
Open challenges for deep reinforcement learning systems are their adaptivity to changing environments and their efficiency w.r.t. computational resources and data. In the application of learning lane-change behavior for autonomous driving, the number of required transitions imposes a bottleneck, since test drivers cannot perform an arbitrary amount of lane changes in the real world. In the off-policy setting, additional information on solving the task can be gained by observing actions from others. While in the classical RL setup this knowledge remains unused, we use other drivers as surrogates to learn the agent's value function more efficiently. We propose \emph{Surrogate Q-learning} that deals with the aforementioned problems and reduces the required driving time drastically. We further propose an efficient implementation based on a permutation equivariant deep neural network architecture of the Q-function to estimate action-values for a variable number of vehicles in sensor range. We evaluate our method in the open traffic simulator SUMO and learn well performing driving policies on the real highD dataset.
\end{abstract}

\section{INTRODUCTION}

For high-level decision making, many autonomous driving systems use pipelines comprised of modular components for perception, localization,  mapping,  motion planning and the decision component itself. In recent years, Deep Reinforcement Learning (DRL) has shown promising results in various domains \cite{DBLP:journals/nature/MnihKSRVBGRFOPB15,DBLP:journals/nature/SilverHMGSDSAPL16,DBLP:journals/jmlr/LevineFDA16}, and in the application of high-level decision making in autonomous driving \cite{branka_rl_highway,Mirchevska2018HighlevelDM,DeepSetQ,DeepSceneQ,kalweit2020inverse,cdqn}. However, there remain a lot of challenges that are limiting w.r.t. the implementation of DRL for such components on real physical systems. Besides others, the following properties have to be ensured:
\begin{enumerate}
    \item \textbf{Generalization:}  The system has to generalize to new situations. Since classical rule-based decision-making modules are limited in their generalization abilities, DRL methods offer a promising alternative for this component due to their ability to learn policies from previous experiences. Nonetheless, generalization has to be improved by using suitable deep network architectures.
    \item \textbf{Adaptivity:} The system has to deal with changing environments and a varying number of traffic participants.
    \item \textbf{Efficiency:} Computational and data-efficiency are restrictive bottlenecks, i.e. the number of required lane-changes of the test vehicle in the training set. While in simulation it is possible to collect data with policies performing as many lane-changes as possible \cite{DeepSetQ,DeepSceneQ}, in real scenarios test drivers cannot perform that many lane-changes. This restriction leads to a tremendous increase in demand of recordings.
\end{enumerate}

\begin{figure}[t!]
    \center
    \includegraphics[width=0.49\textwidth]{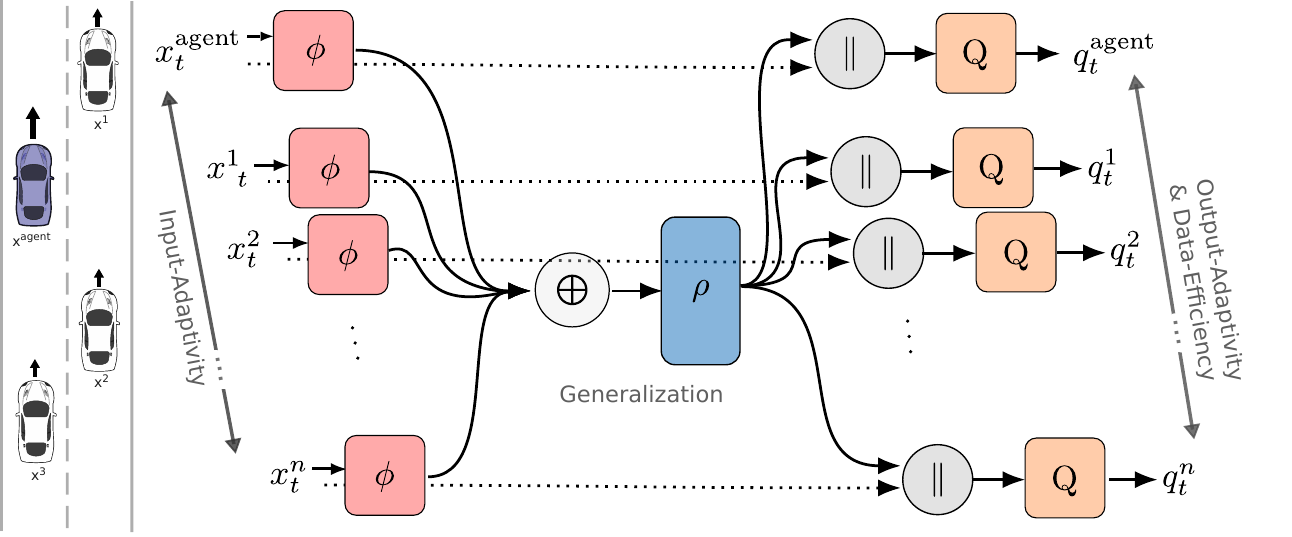}
    \caption{Scheme of the efficient, adaptive and generalizing \SURQ architecture. The modules $\phi$, $\rho$ and $Q$ are fully-connected neural networks and  $\bigoplus$ the \textit{sum} as pooling operation. Element-wise concatenation with input features $x^j_t$ of vehicle $j$ at time step $t$ is denoted by $||$ . Q-values $q_{(\cdot)}$ are computed in parallel for a varying number of vehicles to train the action-value function of the agent efficiently. The optimal policy of the agent can be extracted by $\mathop{\argmax_{a}} q_t^{\text{agent}}$.}
    \label{fig:dsscheme}
\end{figure}
\begin{figure*}[h]
    \center
    \includegraphics[width=0.99\textwidth]{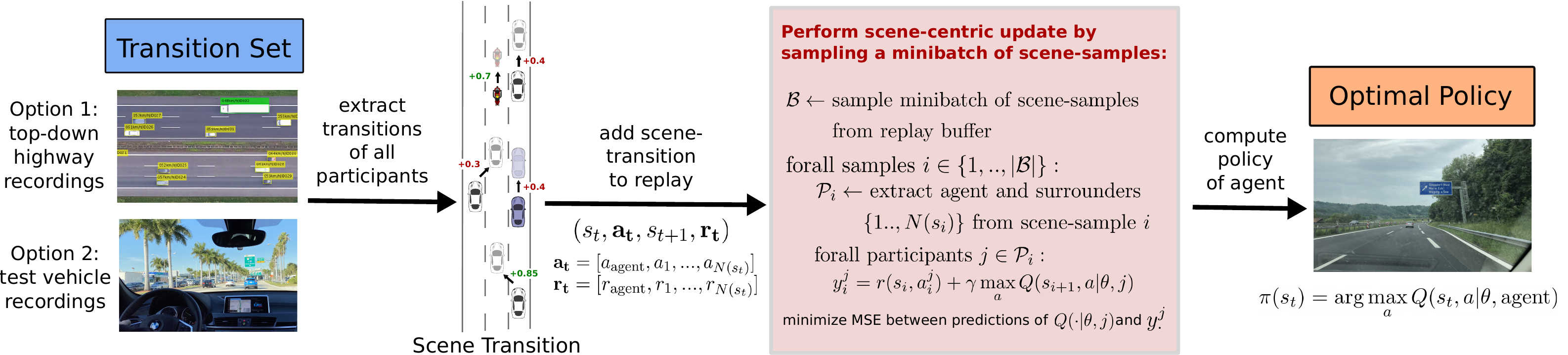}
    \caption{Scheme of Deep Surrogate Q-Learning for the application of autonomous lane-changes. The agent (blue) and surrounding vehicles (white) act in a highway scenario. Exploiting off-policy Q-learning, transitions of all participants can be used to update the agent's Q-function w.r.t the agent's reward function, even if they have different goals and reward functions. The transition set is excluded from top down recordings of German highways in the open-source HighD dataset \cite{highDdataset}.}
    \label{fig:cover}
\end{figure*}

A well-performing flexible architecture for the action-value function that can deal with a variable-sized input list was already proposed in the DeepSet-Q approach \cite{DeepSetQ} for this domain. The architecture tackles already two of the three above mentioned challenges: \textbf{generalization} and \textbf{adaptivity}. The algorithm, employing the formalism of \emph{Deep Sets} \cite{NIPS2017_6931}, is able to deal with a variable number of surrounding vehicles as an extension of DQN to estimate the action-value function for the agent. This approach outperformed DQN agents with fully-connected networks using fixed-sized input representations  \cite{adaptive_behavior_kit,branka_rl_highway,Mirchevska2018HighlevelDM,2018TowardsPH,overtaking_maneuvers_kaushik} or CNNs in combination with occupancy grids \cite{tactical_decision_making,deeptraffic}. DeepSet-Q was able to learn a comfortable and robust driving policy with a transition set of $\sim270$ driving hours in simulation. In \cite{DeepSceneQ}, the input representation based on Deep Sets was extended to deal with graphs in the interaction-aware \mbox{Graph-Q} algorithm.\\ 

For implementation on real physical systems, however, there remains the challenge of building an \textbf{efficient} system while staying adaptive to a changing environment and generalizable to unseen situations. In the domain of autonomous driving, the representation of the environment comprises of an agent, performing actions and collecting data, and a list of surrounding traffic participants, each acting w.r.t. their own respective value-function but within the same high-level action space. The classical RL formulation, only considering the agent's transitions, can be improved by exploiting that recorded transitions contain collections of all vehicles in sensor range, and actions and rewards of all vehicles can be inferred for consecutive time steps using the reward function of the ego-car agent (not the actual, unknown rewards of the other participants). Hence, all transitions of all vehicles can be leveraged via off-policy reinforcement learning (RL) in order to optimize a value-function subject to the reward function of the agent, which we call \textbf{Surrogate Q-learning}. Taking into account actions of all surrounding drivers reduces the required driving time to collect the same amount of lane-changes tremendously. While in the classical RL setup, this additional knowledge about task execution remains unused, in this setting, a policy can even be learned without performing any lane-changes with the test vehicle itself or by recording images from drones or bridges.\\

The key aspect of Surrogate Q-learning is a flexible permutation equivariant deep neural network architecture \cite{NIPS2017_6931} to estimate the action-value function with DQN~\cite{DBLP:journals/nature/MnihKSRVBGRFOPB15} efficiently, as shown in \Cref{fig:dsscheme}.  A \textbf{permutation equivariant} architecture computes for a list of input vectors an equally-sorted and equally-sized list of output vectors. The architecture makes use of transitions of all participants in sensor range in parallel, leading to maximum data- and runtime-efficiency. The network is able to deal with a variable-sized list of inputs, equal to the current number of vehicles in sensor-range and outputs a variable-sized list of output vectors of the same size simultaneously, here corresponding to Q-values for every vehicle in the scene. Since these values are estimated by the same Q-approximation network, the additional transitions help to speed up learning of the action-value estimation of the agent drastically.\\

By taking all actions from all participants in a sample into account for one respective update, Surrogate Q-learning inherently encompasses a unique way of replay sampling. Informed sampling from the replay buffer has been first discussed in \cite {2015arXiv151105952S}, where transitions in the replay-buffer are drawn according to a ranking on the basis of current TD-error. Surrogate Q-learning, on the other hand, implements a sample distribution dependent on the complexity of the scene, i.e. the number of vehicles and takes full advantage of the given sample set without an increase in computation time.\\

Our main contributions are threefold: First, we propose Surrogate Reinforcement Learning which makes use of transitions of all vehicles to generate more actions and rewards by evaluating all other vehicles with the reward function of the agent, as described in \Cref{fig:cover} and formalize the Surrogate Q-learning algorithm. Second, we introduce a novel flexible permutation equivariant network architecture to estimate Q-values for all vehicles in sensor-range efficiently in parallel. Third, we evaluate the performance in the open-source traffic simulator SUMO and furthermore train on the open real highD dataset containing top-down recordings of German highways, which is an important step towards the application on a real system.

\section{REINFORCEMENT LEARNING}
We model the task of learning lane-change behavior in autonomous driving as a Markov Decision Process (MDP), where the agent is acting with policy $\pi$ in a highway environment. In state $s_t \in \mathcal{S}$, the agent applies a discrete action $a_t^{\text{agent}} \in \mathcal{A}$ and reaches a successor state ${s_{t+1}}\sim\mathcal{M}$ according to a transition model $\mathcal{M}$. In every time step $t$, the agent receives a reward $r_{t}^\text{agent}$ for driving comfortably and as close as possible to a desired velocity and tries to maximize the expectation of the discounted long-term return $R(s_t) = \sum_{i>=t} \gamma^{i-t}r_i$, where $\gamma \in [0, 1]$ is the discount factor. The Q-function $Q^\pi(s_t,a_t)=\mathbf{E}_{a_{i>t}\sim\pi}[R(s_t)|a_t]$ represents the value of following a policy $\pi$  after applying action $a_t$. The optimal value-function and optimal policy for a given MDP can then be found via Q-learning \cite{Watkins92q-learning}.

\section{SURROGATE Q-LEARNING}

\label{sec:SurrQ}
We consider tasks in complex and stochastic environments where our agent acts among other participants, each following their own respective policy simultaneously. The set of participants in a scene $\mathbf{s}_t$ at time point $t$ is denoted as $\mathcal{P}(\mathbf{s}_t) = \{\text{agent}, 1, ..., N(\mathbf{s}_t)\}$, where $N(t)$ is equal to the number of surrounders in sensor-range of the agent. We assume that all surrounders have the same action space $\mathcal{A}$ as the agent or actions that are mappable to the agent's action space. Our aim is to optimize the behavior of the agent according to its own reward function, but using additional \emph{surrogate} transition data from other participants, judged by the agent's reward function. This is in contrast to the multi-agent RL setting, where the long-term return of all agents involved is optimized together.
We assume that the agent perceives information about surrounders in the form \mbox{$\mathbf{s_t}=[x_t^\text{agent}, x^1_t, \dots, x^{N(t)}_t]^\top$}, where $x_t^{(\cdot)}$ are feature vectors describing the participants.
In Surrogate Q-learning, we extend the classical RL setting by considering a vector of actions \mbox{$\mathbf{a}_t = [a_t^{\text{agent}}, a^1_t, a^2_t, ..., a^{N(t)}_t]^\top$} and rewards \mbox{$\mathbf{r}_t = [r_t^{\text{agent}}, r^1_t, r^2_t, ..., r^{N(t)}_t]^\top$} for all participants, as demonstrated in \Cref{fig:cover}. The scalar rewards are computed according to the ego-car agent's reward function, which we define as a global reward function in a more general fashion than in the classical RL setting by $r^{\text{agent}}: \mathcal{S} \times \mathcal{A} \times \mathcal{S} \times \mathcal{P}  \rightarrow  \mathbb{R}$, where $r_t
^p = r^{\text{agent}}\left(\textbf{s}_t, a_t^p, \textbf{s}_{t+1} | p\right)$ is evaluating the action of an arbitrary participant $p$ according to the agent's reward function. This is possible whenever the agent can detect or infer actions of its surrounders. Based on the agent's reward function, we can estimate the long-term value $Q(\textbf{s}_t, a_t^p|\theta, p)$ for the action executed by participant $p\in \mathcal{P}(\textbf{s}_t)$. Weights $\theta$ are parameters of the value-function approximation. The final policy of the agent can be extracted by:
$$\pi(\mathbf{s}_t) = \argmax_{a \in \mathcal{A}} Q(\mathbf{s}_t, a|\theta, \text{agent}).$$

In Deep Surrogate Q-learning, we use DQN to learn the optimal policy $\pi$ and fill a replay buffer  $\mathcal{R}$ with scene-transitions $\kappa_i =  (\mathbf{s}_i, \mathbf{a}_i, \mathbf{s}_{i+1}, \mathbf{r_i})$.  Naively, a new replay buffer $\mathcal{R}'$ could be created by $\mathcal{R}' = \bigcup_{\kappa_i \in \mathcal{R}} \beta^\mathcal{R}(\kappa_i)$ with a projection function $\beta^\mathcal{R}(\kappa_i) = \{(\mathbf{s}_i, a_i^p, \mathbf{s}_{i+1}, r_i^p) | \  \forall p \in \mathcal{P}(\mathbf{s}_i)\}$ and minibatches could be sampled uniformly from $\mathcal{R}'$ to update the Q-values. 
Instead, Surrogate Q-learning implements a novel experience replay sampling technique, which focuses on the complexity of a scene. To implement it efficiently, we introduce a novel permutation equivariant architecture capable of estimating multiple action-values at once, as detailed below.\\

In this setup, we define the complexity of a scene to be proportional to the number of surrounders. Intuitively, navigation in scenes with many close surrounders can be much more complicated than in empty spaces.
From every scene-sample $\kappa_i$ in a minibatch $\mathcal{B}$ of size $m$, a further batch of samples can be generated via a projection function $\beta^\mathcal{B}$, which extracts a batch of transitions of all participants in the scene: $$\beta^\mathcal{B}(\kappa_i) = \{(\mathbf{s}_i, a_i^p, \mathbf{s}_{i+1}, r_i^p) | \  \forall p \in \mathcal{P}(\mathbf{s}_i)\}.$$ 
 
 To update the action-value function, we then create a batch of \emph{virtual} samples by concatenation of the projected batches:
$$\mathcal{B}^{\text{virtual}} = \bigcup_{1\le i \le m} \beta^\mathcal{B}(\kappa_i).$$ 

The size of the virtual batch can then be computed by $|\mathcal{B}^{\text{virtual}}| = \sum_{1 \le i \le m} N(s_i)$. We sample minibatches of $\mathcal{R}$ uniformly with $\kappa_i \sim \mathcal{R}$, which means that the proportion of a scene in the virtual batch is larger the more surrounders are in the given scene. The optimization of Surrogate-Q is then formalized as: 
$$\min_\theta \mathbb{E}_{\mathcal{B}^\text{virtual} \sim \mathcal{R}} [ \sum_{(\mathbf{s}_i, a_i^p, \mathbf{s}_{i+1}, r_i^p) \in \mathcal{B}^\text{virtual} } (y_i^p - Q(\textbf{s}_i, a_i^p | \theta, p))^2],$$
with targets $y_i^p = r_i^p + \gamma \max_{a \in \mathcal{A}} Q'(\textbf{s}_{i+1}, a|\theta', p)$, where $Q'$ is a target network with parameters $\theta'$.\\

While execution of the learned policy is fast and efficient, computation requires multiple forward-passes per sample for classical deep neural network architectures, such as fully-connected neural networks or DeepSets. Depending on the number of surrounders, the  runtime lies in $O(b \times m \times n)$ with $n = \max_{\textbf{s}_i \in \mathcal{R}}  N(\mathbf{s}_i)$ for replay buffer $\mathcal{R}$ and the number of sampled batches $b$. To reduce runtime, we describe an efficient permutation equivariant architecture in \Cref{sec:permequi}, which calculates Q-values for all participants in a scene in parallel, resulting in a runtime of $O(b \times m)$.

\section{EFFICIENT SURROGATE Q-LEARNING VIA PERMUTATION EQUIVARIANT Q-NETWORKS}
\label{sec:permequi}
We use a permutation equivariant architecture to estimate the action-value function of the agent efficiently. A permutation equivariant architecture can  more formally be defined as a function $f : X^M \rightarrow Y^M$ that keeps the input permutation for the output, i.e. for any permutation $\pi$: $f([x_{\pi(1)}, ..., x_{\pi(M)}]) = [f_{\pi(1)}(x), ..., f_{\pi(M)}(x)]$. Similar as in \cite{DeepSetQ,DeepSceneQ}, the architecture is able to deal with a variable number of input elements. The action-value function is represented with a deep neural network $Q_{\text{SUR}}(\cdot, \cdot|\theta, \mathcal{P}(\cdot)))$, parameterized by weights $\theta^{Q_{\text{SUR}}}$, and optimized via DQN. The network, consisting of three network modules $(\phi,\rho,Q)$,  outputs a vector of action values for all participants $\mathcal{P}$ of the current scene, where architectural choices follow the derivations of the permutation equivariant case in \cite{NIPS2017_6931}. The input layers are built of two neural networks $\phi$ and $\rho$, which are components of the Deep Sets \cite{NIPS2017_6931}, similarly used as in the work of \cite{DeepSetQ} to deal with a variable input. The representation of the input scene $\mathbf{s}_t$ is computed by: 
 $$\Psi(\mathbf{s}_t) = \rho\left(\sum_{x^j_t\in \mathbf{s_t}}\phi(x_t^j)\right),$$ 
 resulting in a permutation invariant representation of the scene at time step $t$. In this work, we combine this scene-representation, concatenate it with features of the participants, and feed the resulting latent vector to the $Q$ module. Every output is the Q-value corresponding to one vehicle in the scene. More formally, the list of Q-values $\mathbf{q}_t = [q_t^\text{agent}, q_t^1, ..., q_t^{N(\mathbf{s}_t)}]$ is calculated as:
 
 $$\mathbf{q}_t = Q_{\text{SUR}}\left(\bm s_t, \mathbf{a}_t|\theta, \mathcal{P}(\mathbf{s}_t)\right), $$
$$\text{where } q^p_t =  Q(\epsilon(\Psi(\bm s_t), x_t^p), a_t^p|\theta, p),$$ 
for every vehicle $p\in \{\text{agent}, 1, ..., N(\mathbf{s}_t) \}$, as demonstrated in \Cref{fig:dsscheme}. We choose $\epsilon(\Psi(\bm s_t), x_t^p) = [\Psi(\bm s_t)|| x^p_t]$ to be the concatenation of participant features $x_t^p$ and the combined scene representation, leading to a prediction of the action-value for every participant. During runtime, the optimal policy of the agent can be extracted by:
$$\pi(\mathbf{s}_t) = \mathop{\argmax_{a}} q_t^{\text{agent}}.$$

\begin{algorithm}[t]
    \SetAlgoLined
    \DontPrintSemicolon
    initialize $Q_{\text{SUR}}=(\phi,\rho,Q)$ and ${Q_{\text{SUR}}}'=(\phi',\rho',Q')$\\
    set replay buffer $\mathcal{R}$\\

    \For{\text{optimization step} o=1,2,\dots}{
        get minibatch $\mathcal{B}^\text{virtual}$ from $\mathcal{R}$\\
        \ForEach{$(\bm s_i,\bm a_i, \bm s_{i+1}, \bm r_{i})$ \text{ in } $\mathcal{B}^\text{virtual}$ }{
            \ForEach{\text{vehicle }$x_{i+1}^j$ in $\bm s_{i+1}$}{
                $({\phi'}_{i+1})^j=\phi'\left(x_{i+1}^j\right)$\\
            }
            $\Psi'(\bm s_{i+1})=\rho'\left(\sum\limits_j({\phi'}_{i+1})^j\right)$\\
             \ForEach{\text{vehicle }$x_{i+1}^j$ in $\bm s_{i+1}$}{
                $y^j_i=r_i^j + \gamma \max_{a} Q'(\epsilon ({\Psi'(\bm s_{i+1})}, x^j_{i+1}), a)$\\
            }
        }
        perform a gradient step on loss:
        $ \frac{1}{m}\sum\limits_i \sum\limits_{j} \left(y_i^j - Q^j_{\text{SUR}}(\bm s_i, a_i^j)\right)^2$\\
        update target network by $\theta^{{Q_{\text{SUR}}}'}\leftarrow\tau\theta^{Q_{\text{SUR}}} + (1-\tau)\theta^{{Q_{\text{SUR}}}'}$
        
    \For{\text{execution step} e=1,2,\dots}{
        get current state $\bm s_t$ from environment\\
        $\text{apply } a^{\text{agent}}= \mathop{\argmax_{a}} Q^{\text{agent}}_{\text{SUR}}(\bm s_t, a)$\\
    }
    }
    \vspace{0.2475cm}
    \caption{Deep Surrogate Q-Learning via permutation equivariant Q-Networks}
    \label{alg:setq}
\end{algorithm}

The scheme of the network architecture is shown in \Cref{fig:dsscheme}. The Q-function is trained on virtual minibatches $\mathcal{B}^\text{virtual}$ as described in \Cref{alg:setq}.

\section{EXPERIMENTS}

We formalize the task of performing autonomous lane changes as an MDP. The state space consists of relative distances, relative velocities and relative lanes for all vehicles within the maximum sensor range of the vehicle. As action space, we consider a set of discrete actions $\mathcal{A}$ in lateral direction, including \emph{keep lane}, \emph{perform left lane-change} and \emph{perform right lane-change}. Acceleration is handled by a low-level execution layer with model-based control of acceleration to guarantee comfort and safety. We use RL to optimize the long-term return, for which model-based approaches are limited in this domain. Collision avoidance in lateral direction and maintaining safe distance in longitudinal direction are controlled by an integrated safety module, analogously to
\cite{deeptraffic}, \cite{Mirchevska2018HighlevelDM,DeepSetQ}. For unsafe actions, the agent keeps the lane. We define the reward function as: 
$$r^\text{agent}(\mathbf{s}_{t}, a^p, \mathbf{s}_{t+1}| p) = 1 - \frac{|v_t^p- v_{\text{desired}}|}{v_{\text{desired}}} - p_{\text{lc}}(a^p), $$

where $v_t^p$ is the current velocity of participant $p$ and $p_{\text{lc}} = 0.01$ if action $a^p$ is a lane change and $0$ otherwise. The weight was chosen empirically in preliminary experiments. Lane-changes of surrounding vehicles can be detected by a change of lane index in two consecutive time steps. To ensure the same number of vehicles in two successive states of a transition, during training, dummy vehicles are added in case they only appear in one of the states because they leave or enter sensor range. This is necessary for the calculation of the Q-values with the permutation equivariant network architecture. In the execution phase this is not relevant.

\subsection{Comparative Analysis}

We compare Deep \SURQ to the rule-based controller of SUMO with lane change model \textit{LC2013} and to the state-of-the-art DeepSet-Q algorithm. Since it has already been shown that a Deep Set representation is outperforming other methods using input representations such as fully-connected, recurrent or convolutional neural network modules for this specific application at hand, we omit to include these other baselines in this work and refer to \cite{DeepSetQ} for an extensive comparison. In DeepSet-Q, the features of all surrounding vehicles are projected to a latent space by the network module $\phi$ in a similar manner as in Surrogate-Q. The combined representation of all surrounding vehicles is computed by $ \Psi( \bm s_t \backslash x_t^{\text{agent}})= \rho\left(\sum_{x\in \bm s_t \backslash x_t^{\text{agent}}}\phi(x) \right)$. Static features describing the agents state $x^{\text{agent}}$ are fed directly to the $Q$-module, and the Q-values are computed by \mbox{$Q_{\mathcal{DS}} = Q(\Psi(\bm s_t \backslash x_t^{\text{agent}}) || x^{\text{agent}},a)$}, where $||$ denotes a concatenation of two vectors and $a$ is the action of the agent. The updates are performed for every transition $i$ in a minibatch of size $m$ with  $y_i = r_i+\gamma\max_a Q'(\Psi'(\bm s_{i+1}  \backslash x_{i+1}^{\text{agent}}) ||  x_{i+1}^{\text{agent}}, a)$ using a slowly updated target network $Q'$. To update parameters we perform a gradient step on the loss $\frac{1}{m}\sum_i\left(Q_{\mathcal{DS}}(\bm s_i, a_i)-y_i\right)^2$. If not denoted differently, every network is trained with a batch size of $64$ for $2.5\cdot10^6$ gradient steps and optimized by Adam \cite{DBLP:journals/corr/KingmaB14} using a learning rate of $10^{-4}$. Rectified Linear Units (ReLu) are used as activation function in all hidden layers of all architectures. We update target networks by Polyak averaging. To prevent the predictions from overestimation, we further apply Clipped Double-Q learning \cite{DBLP:conf/icml/FujimotoHM18}. Target networks are updated with a step-size of $\tau=10^{-4}$. The architectures were optimized using Random Search with the settings shown in \cite{DeepSetQ}. The neural network architectures are shown in \Cref{tab:networks1}.

\begin{table}[t] 
\centering
\caption{Network architectures of DeepSet-Q and \SURQ. FC($\cdot$) denote fully-connected layers.}
 \begin{tabular}{c|c}
       \toprule
       DeepSet-Q & \textbf{\SURQ} \\
        \midrule
        Input($B \times \text{seq len} \times 3$) &  Input($B \times \text{seq len} \times 6$)  \\
        \midrule
         $\phi$: FC($20$), FC($80$) &  $\phi$: FC($20$), FC($80$)\\
              sum($\cdot$) &  sum($\cdot$) \\
        $\rho$: FC($80$), FC($20$) &  FC($80$), FC($80$)\\
        concat($\cdot$, Input($B \times 3$)) &  $\forall x_j$:  concat($\cdot$, $x_j$)\\
       $Q$:  FC(100), FC(100)&  $Q$:  FC($80$), FC($80$)\\
       \midrule
       Output($B \times 3$) & Output($B \times \text{seq len} \times 3$)\\
      
         \bottomrule
    \end{tabular}
        \label{tab:networks1}
\end{table}

\subsection{Simulation}
We use the open-source SUMO traffic simulator \cite{sumo} and perform experiments on a $1000\,$m circular highway with three lanes, with the same settings as proposed in \cite{DeepSetQ}. To create a realistic traffic flow, vehicles with different driving behaviors are randomly positioned across the highway. The behaviors were varied by changing the SUMO parameters maxSpeed, lcSpeedGain and lcCooperative. For training, all datasets were collected on scenarios with a random number of 30 to 90 vehicles. Agents are evaluated for every training run on different scenarios to smooth out the high variance in the stochastic and unpredictable highway environment. The number of vehicles varies from 30 to 90 cars, and for each fixed number of vehicles, 20 scenarios with different a priori randomly sampled positions and driving behaviours for each vehicle are evaluated for every approach. In total, each agent is evaluated on the same 260 scenarios. If not denoted differently, we use 10 training runs for all agents and show the mean performance and standard deviation for the scenarios as described above. The SUMO settings of the experiments are as follows: Sensor range was set to $\SI{80}{\metre}$, time step length of SUMO to $\SI{0.5}{\second}$, action step length and lane change duration $\SI{2}{\second}$. Acceleration and deceleration of all vehicles were $\SI{2.6}{\metre\per\second\squared}$ and $\SI{4.5}{\metre\per\second\squared}$, the vehicle length $\SI{4.5}{\metre}$, minimum gap  $\SI{2}{\metre}$ and desired time headway \mbox{$\tau=\SI{0.5}{\second}$}. As lane change controller \textit{LC2013} with \mbox{$\text{lcKeepRight}=0.0$} was used.

\subsection{Real Data}

Additionally, we trained on transitions extracted from the open-source HighD traffic dataset \cite{highDdataset}. The dataset consists of 61 top-down recordings of German highways, resulting in a total of 147 driving hours. The recordings are preprocessed and contain lists of vehicle ids, velocities, lane ids and positions. We use all tracks with 3 lanes. Additionally, we filter a time span of \SI{5}{\second} before and after all lane changes in the dataset with a step size of \SI{2}{\second}, leading to a consecutive chain of 5 time steps and in total a replay buffer of $\sim18.000$ transitions with $\sim4000$ lane changes. A visualization of the dataset is shown in \Cref{fig:cover} (left).

\begin{figure}[t]
    \center
    \includegraphics[width=0.45\textwidth]{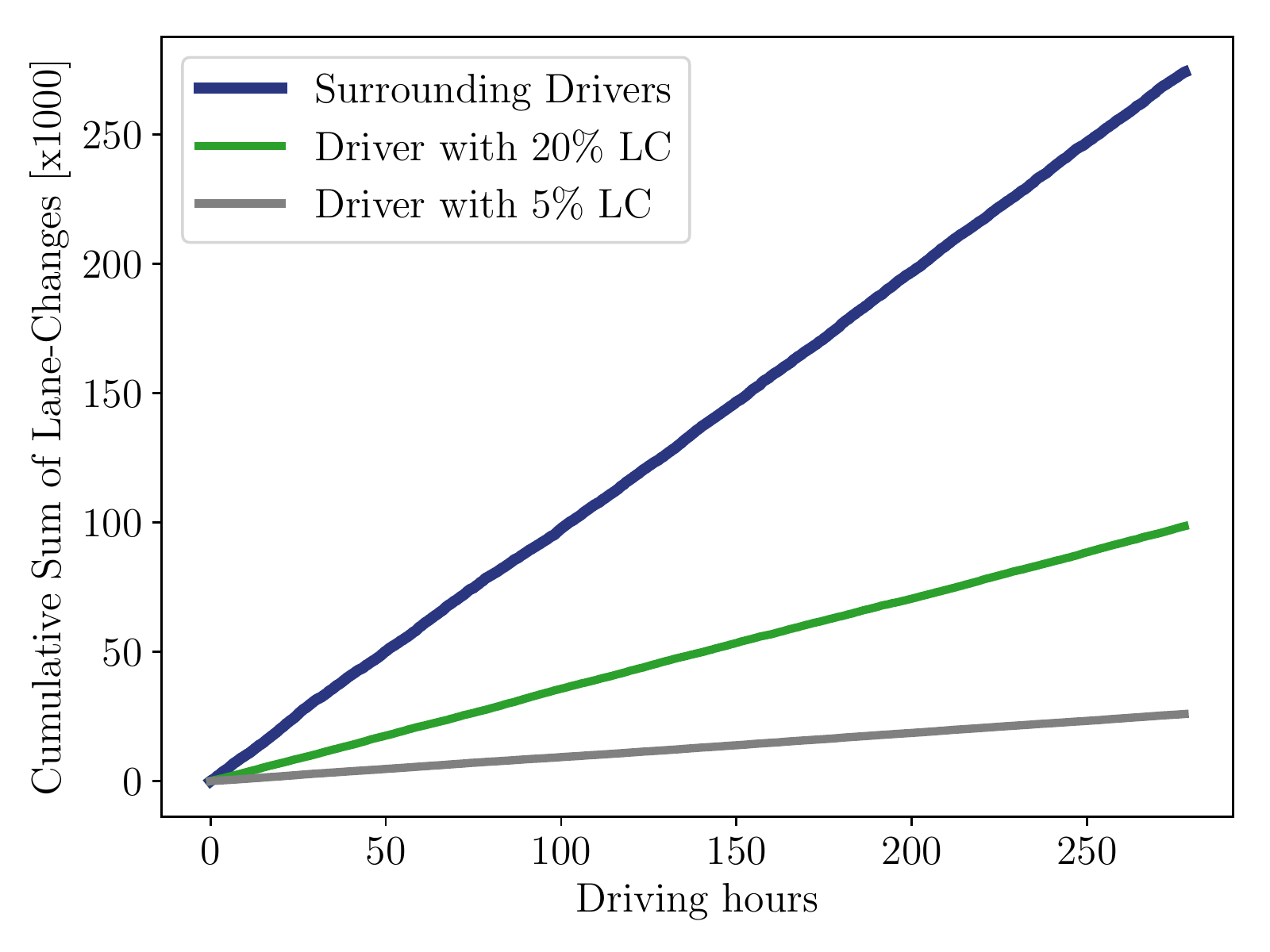}
    \caption{Cumulative sum of lane-changes per driving hours, for transitions collected by different driver models on a highway in the open-source traffic simulator SUMO \cite{sumo}.}
    \label{fig:scenario}
\end{figure}

\begin{figure*}[t]
    \center
    \begin{subfigure}[c]{0.45\textwidth}
    \includegraphics[width=\textwidth]{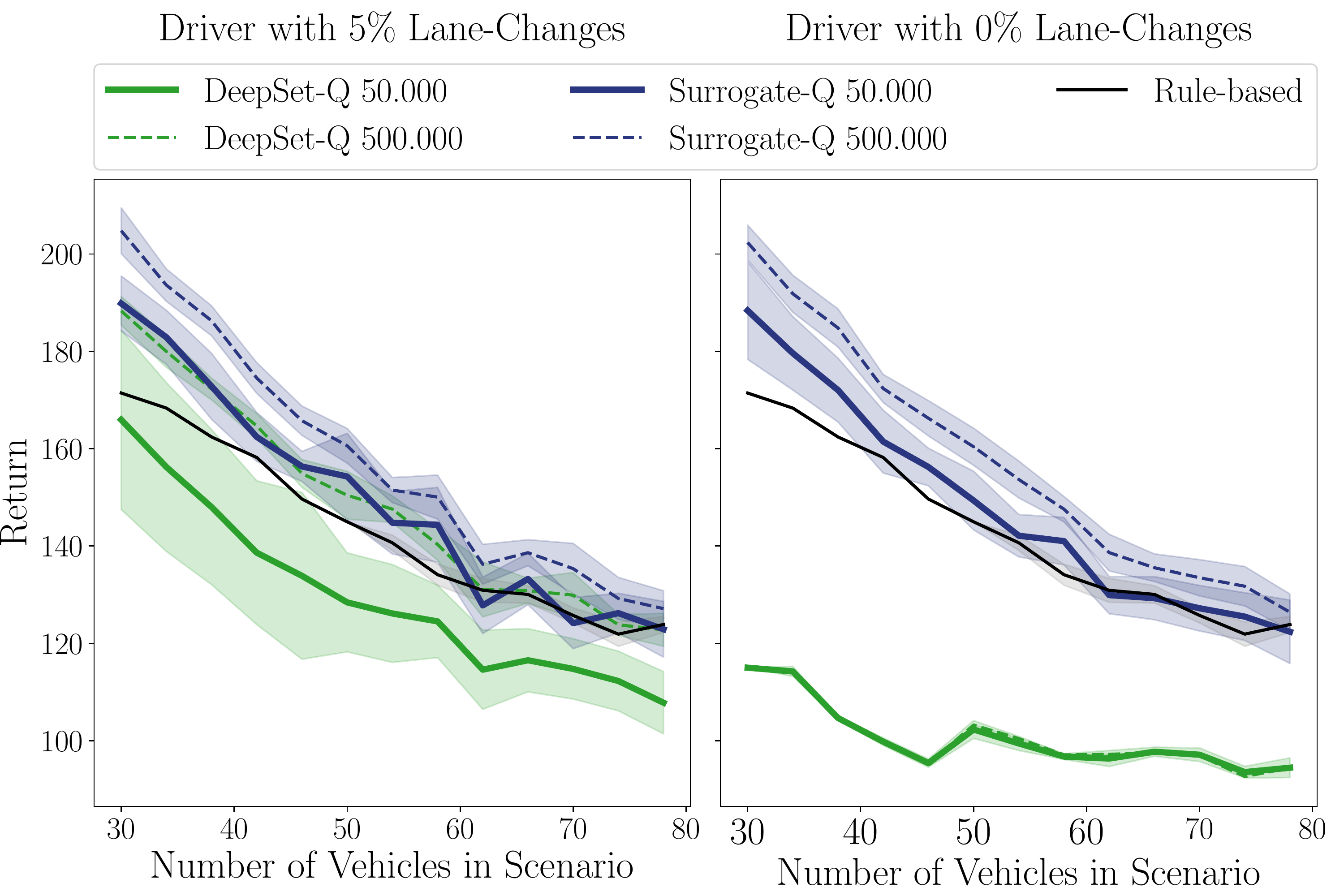}
    \subcaption{}
    \label{fig:res1a}
\end{subfigure}
    \begin{subfigure}[c]{0.25\textwidth}
    \vspace*{0.4cm}
    \includegraphics[width=\textwidth]{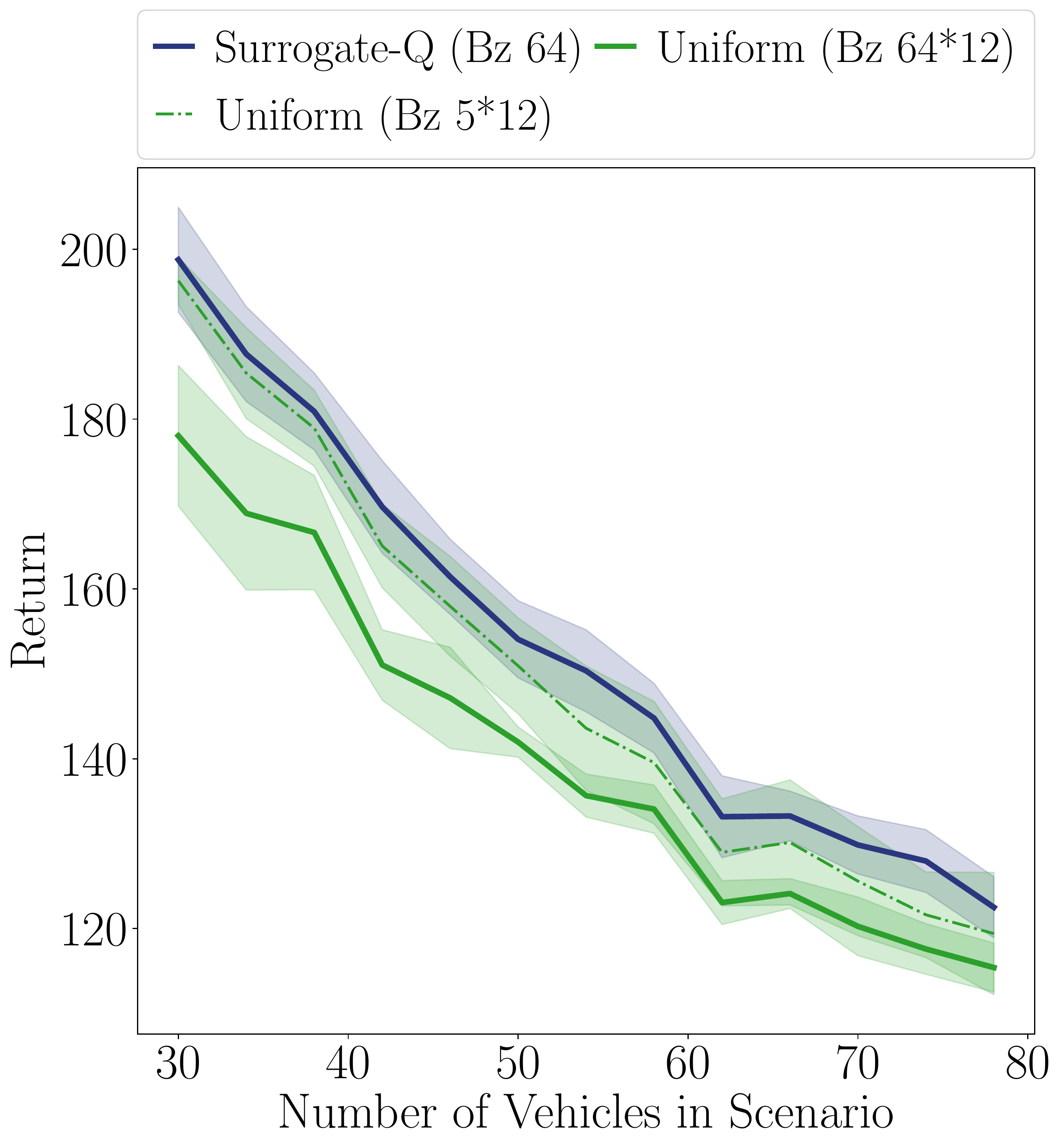}
    \subcaption{}
    \label{fig:res1b}
\end{subfigure}
    \begin{subfigure}[c]{0.25\textwidth}
    \vspace*{0.4cm}
    \includegraphics[width=\textwidth]{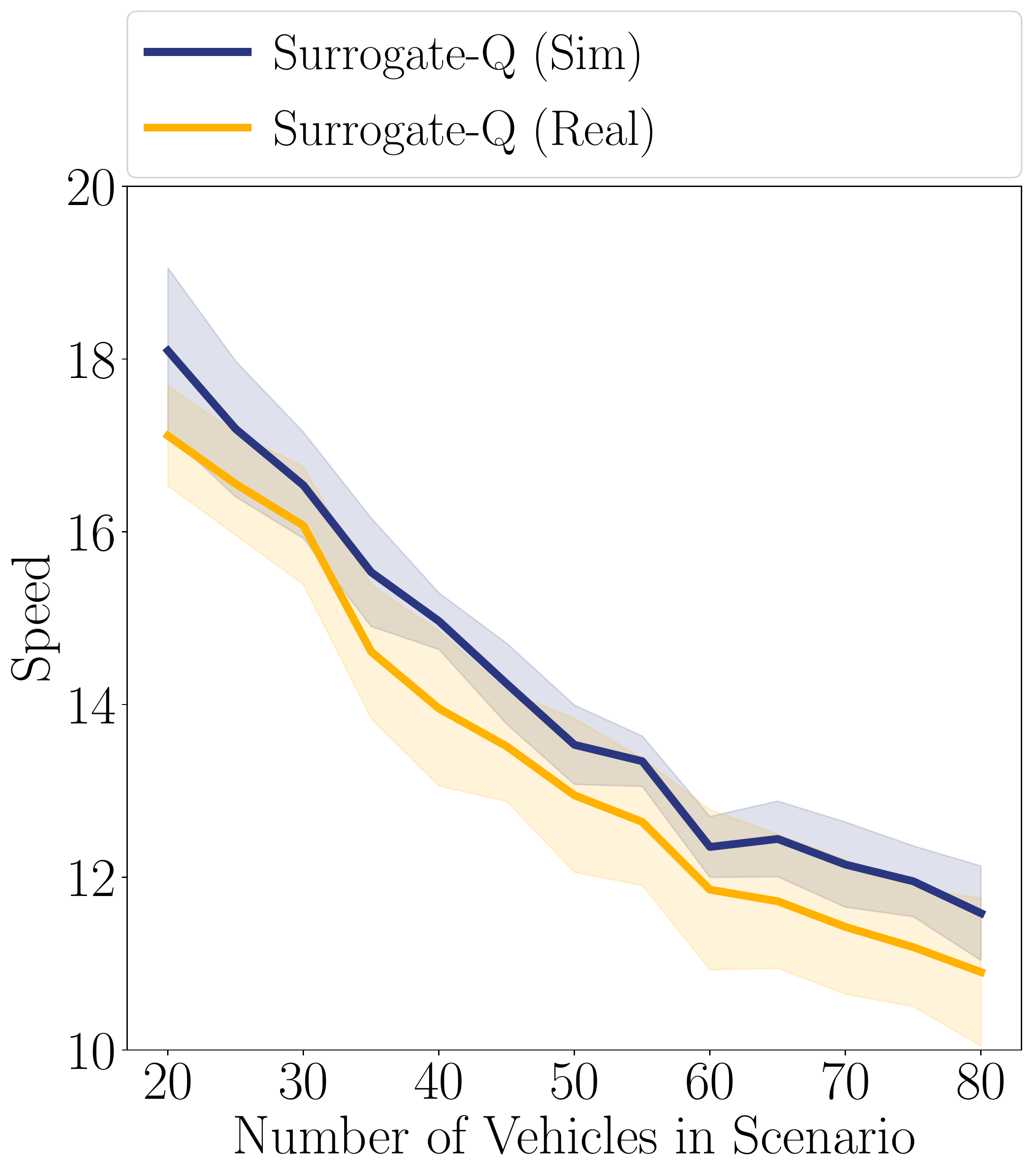}
    \subcaption{}
    \label{fig:res1c}
\end{subfigure}
  
    \caption{Mean performance and standard deviation of over 10 training runs for (a) \SURQ in the highway scenario in SUMO, collected with a driver performing $5\%$ lane-changes (left) and a driver performing no lane-changes (right). Results are shown for different numbers of transitions in the training set, indicated by solid and dashed lines. The number of 500.000 transitions corresponds to $\sim 280$ driving hours and 50.000 transitions to $\sim 28$ driving hours, respectively. (b) \SURQ compared to DeepSet-Q with uniform sampling for varying batch sizes (denoted by Bz). (c) \SURQ trained on the HighD dataset (Real) and trained in simulation (Sim) on 50.000 training samples.}
    \label{fig:res1}
\end{figure*}

\section{RESULTS}
\label{sec:results}

First, we study the required driving time to collect a certain number of lane-changes, considering transitions from test-drivers performing 5\% and 20\% lane-changes and transitions collected from all drivers in sensor-range. \Cref{fig:scenario} shows the cumulative sum of lane-changes for the two test drivers and the total sum of lane-changes of all surrounding vehicles in the \SI{1}{\km} highway scenario as proposed in \cite{DeepSetQ}, with 30 up to 90 vehicles per scenario and with $12$ surrounding vehicles around the agent on average. A test driver performing as many lane-changes as possible (20\% lane-changes) is compared to a more realistic driver model\footnote{Driver model of the SUMO traffic simulation using the \emph{Krauss} car following model in combination with \emph{LC2013} controller.} performing only 5\% lane-changes. The latter requires a much higher amount of driving hours in order to collect the same amount of lane-changes.\\

We compared the Surrogate-Q agent with a DeepSet-Q agent in the same highway scenario in SUMO. Results are shown in \Cref{fig:res1a}. As additional measure of performance, we choose the sum of mean return over all scenarios and provide significance tests using  Welch’s t-test. For a large amount of 500.000 transitions ($\sim 280$ driving hours) collected by a driver with  $5\%$ lane-changes, \SURQ is outperforming Deepset-Q with a p-value of $< 0.001$.  Decreasing the amount of transitions by 1/10 leads to a high decrease of performance for DeepSet-Q. For 50.000 transitions ($\sim 28$ driving hours), DeepSet-Q shows significantly lower performance than Surrogate-Q with a p-value of $< 0.001$. Additionally, \SURQ is outperforming the rule-based controller for light traffic up to 60 vehicles (p-value $<$ 0.01) when trained on only 28 driving hours. Please note that differences between the approaches get smaller the more vehicles are on track because of maneuvering limitations in dense traffic. \SURQ is even able to show similar performance without performing any lane-changes with the test vehicle at all, only by observing the transitions of other vehicles. This drastically simplifies data-collection, since such a transition set could be collected by setting up a camera on top of a bridge above a highway. The DeepSet-Q algorithm is not able to learn to perform lane-changes without the test vehicle itself performing lane-changes in the dataset. Thus, it learns to stay on its lane and achieves a much lower velocity than Surrogate-Q.\\

The optimization of the action-value function for all vehicles in a scene \textit{virtually} increases the batch size by augmenting the minibatch with one distinct imaginary transition for each of them. In order to evaluate the added value of updating w.r.t. different positions in the same scene in parallel, we compare our approach to a DeepSet-Q agent with an analogously increased replay buffer and batch size, but with an underlying uniform sampling distribution. This results in an agent updated with the same number of samples per minibatch, with the only difference being the sampling distribution. Since there is a variable batch size depending on the number of vehicles in Surrogate-Q, we multiply the default batch size by 12 which corresponds to the average number of surrounding cars. The results of the adaption of the batch-size for DeepSet-Q in order to achieve the same number of updates per minibatch than in \SURQ is shown in \Cref{fig:res1b} for a transition set of size 100.000. Both approaches are trained for the same amount of $1.25\cdot10^6$ gradient steps\footnote{Due to computational complexity of DeepSet-Q with a batch-size of 768, we evaluate only 5 training runs for this setting.}. \SURQ is outperforming the uniform sampling technique with a small batch-size of $5 \cdot 12$, showing a p-value of $< 0.05$ and with a large batch-size of $64 \cdot 12$ with a p-value of $<$ 0.001. A higher batch size, which leads to the same amount of updates per batch as for Surrogate-Q, shows a significant performance decrease. This emphasizes the advantage of the replay sampling in Surrogate Q-learning.  Our findings suggest that Surrogate Q-learning via a permutation equivariant architecture leads to a more consistent gradient, since the TD-errors 
are normalized w.r.t. all predictions for the different positions in the scene while keeping the i.i.d. assumption of stochastic gradient descent by sampling uniformly from the replay buffer. Besides, Surrogate Q-learning offers the possibility to evaluate several actions at once because all vehicles are represented by the same features and all execute the same actions. The evaluation of multiple actions simply leads to the agent having access to more knowledge (w.r.t. one specific state) in shorter time, which is an advantage compared to DeepSet-Q. Additionally, the training of the permutation equivariant architecture is tremendously more efficient. The training of DeepSet-Q with a batch size of 768 takes 6 days, the training of \SURQ only 12 hours on a Titan Black GPU for the same amount of updates. The performance of the agent trained on the real dataset consisting of approximately 18.000 transitions is shown in \Cref{fig:res1c}. Despite mismatches between simulation and the real recordings, \SURQ trained on HighD shows a comparable performance to the agent trained on 50.000 transitions in simulation when evaluated in SUMO.

\section{CONCLUSION}

We introduced a novel deep reinforcement learning algorithm, which can efficiently be implemented via a flexible permutation equivariant neural network architecture. Exploiting off-policy learning, the algorithm takes transitions of all vehicles in sensor range into account by considering a global reward function. \SURQ is extremely efficient in terms of the required number of transitions and with respect to training runtime. Due to the novel architecture of the Q-network, the agent can efficiently exploit all useful information in a given transition set, which alleviates the problem of data collection with a test vehicle significantly. Data can be collected by recordings from top-down views of highways (e.g. from bridges or drones), which simplifies the pipeline in training autonomous vehicles tremendously. Additionally, we successfully showed that Surrogate-Q can be trained on real data. 

\bibliographystyle{unsrt}
\balance
\bibliography{root}

\end{document}